\documentclass[conference]{IEEEtran}
\usepackage{cite}
\usepackage{amsmath,amssymb,amsfonts}
\usepackage{algorithmic}
\usepackage{graphicx}
\usepackage{textcomp}
\usepackage{xcolor}
\usepackage{booktabs}
\usepackage{soul}

\begin{document}

\title{Towards an AI Coach to Infer Team Mental\\ Model Alignment in Healthcare}
\author{
\IEEEauthorblockN{Sangwon Seo}
\IEEEauthorblockA{\textit{Department of Computer Science}\\
\textit{Rice University}\\
Houston, TX, USA\\
\texttt{sangwon.seo@rice.edu}}\\
\IEEEauthorblockN{Julie A. Shah}
\IEEEauthorblockA{\textit{Massachusetts Institute of Technology}\\
Cambridge, MA, USA\\
\texttt{julie\_a\_shah@csail.mit.edu}}
\and
\IEEEauthorblockN{Lauren R. Kennedy-Metz }
\IEEEauthorblockA{\textit{Harvard Medical School and}\\
\textit{U.S. Dept. of Veterans Affairs}\\
Boston, MA, USA\\
\texttt{Lauren.Kennedy-Metz@va.gov}}\\
\IEEEauthorblockN{Roger D. Dias}
\IEEEauthorblockA{
\textit{Brigham and Women's Hospital}\\
\textit{Harvard Medical School}\\
Boston, MA, USA\\
\texttt{rdias@bwh.harvard.edu}}
\and
\IEEEauthorblockN{Marco A. Zenati}
\IEEEauthorblockA{\textit{Harvard Medical School and}\\
\textit{U.S. Dept. of Veterans Affairs}\\
Boston, MA, USA\\
\texttt{marco\_zenati@hms.harvard.edu}}\\               
\IEEEauthorblockN{Vaibhav V. Unhelkar}
\IEEEauthorblockA{\textit{Department of Computer Science}\\
\textit{Rice University}\\
Houston, TX, USA\\
\texttt{unhelkar@rice.edu}}
}

\maketitle

\begin{abstract}
Shared mental models are critical to team success; however, in practice, team members may have misaligned models due to a variety of factors.
In safety-critical domains (e.g., aviation, healthcare), lack of shared mental models can lead to preventable errors and harm.
Towards the goal of mitigating such preventable errors, here, we present a Bayesian approach to infer misalignment in team members’ mental models during complex healthcare task execution.
As an exemplary application, we demonstrate our approach using two simulated team-based scenarios, derived from actual teamwork in cardiac surgery.
In these simulated experiments, our approach inferred model misalignment with over 75\% recall, thereby providing a building block for enabling computer-assisted interventions to augment human cognition in the operating room and improve teamwork.
\end{abstract}

\begin{IEEEkeywords}
teamwork, surgical data science, cardiac surgery, Bayesian inference, patient safety, artificial intelligence
\end{IEEEkeywords}

\section{Introduction}

Alignment of mental models among the members of a team is critical to achieving effective teamwork.
In absence of shared understanding \cite{converse1993shared} about the goals, plans and context of the team, teamwork often results in preventable errors \cite{baker2005medical, kolander2019flight}.
For instance, lack of shared mental models between members of a flight crew can result in aviation accidents \cite{kolander2019flight}.
Similarly, misalignment between members of a surgical team can lead to preventable harm and adverse events \cite{wahr2013patient}.
Teams can adopt best practices (such as team training and debriefing) to improve their mental model sharing \cite{van2011team, salas2013developing, neily2010association}.
Nevertheless, the possibility of preventable error persists due to the impact of execution-time factors, such as high workload, surgical flow disruptions or fatigue \cite{zenati2019first}.

\begin{figure}
\centerline{%
    \includegraphics[width=0.65\columnwidth]{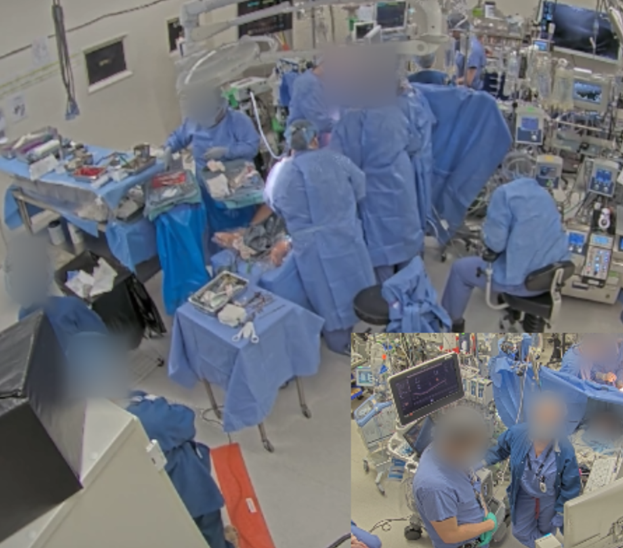}
}
\caption{Teamwork in the cardiac operating room: Surgery being performed by a team of surgeons, anesthesiologists, perfusionists, and nurses.}
\label{fig:teamwork-or}
\end{figure}

Informed by the challenges of teamwork in the cardiac operating room (Fig.~\ref{fig:teamwork-or}), our goal is to mitigate preventable errors of human teams performing goal-oriented and time-critical tasks.
Team assessments, through questionnaires and metrics of evaluating team fluency, provide one avenue for teams to improve shared understanding and teamwork \cite{britton2017assessing, macalpine2017evaluating, dias2021dissecting}.
However, due to their inherent post-hoc nature, it is difficult to utilize these assessments during time-critical tasks and alleviate preventable errors arising due to execution-time factors.
Moreover, while collaborating, it is not easy for team members to self-assess their teamwork, due to distributed cognition and the partially observable nature of collaboration.

We posit that augmenting post-hoc assessments with on-the-fly interventions can help mitigate preventable errors caused due to execution-time factors.
We envision a digital team member (\emph{AI Coach}) that can assess and improve teamwork by monitoring the team members during their collaborative task execution and providing timely interventions.
Due to the advances in surgical data science, sensing hardware and software \cite{maier2020surgical, kennedy2020sensors, kennedy2020computer,  kim2018intelligent}, the opportunity is ripe to develop such an execution-time tool.
However, several challenging problems need to be resolved to realize its vision.
For instance, based on the sensed information, the AI Coach would need to infer whether the mental models (a latent quantity) are aligned.
Similarly, it would need to identify when to provide interventions to improve teamwork.
In this work, we describe our research towards realizing one such capability: namely, execution-time inference of the alignment of team member’s mental models.

We focus on teams with fixed membership and performing sequential collaborative tasks in stochastic environments.
Specifically, we consider collaborative tasks that can be represented as a partially observable variant of Markov decision processes (MDPs), where the latent variables represent the team’s preference (or, equivalently, mental model) regarding collaborative task execution \cite{oliehoek2016concise}.
In the ideal case each team member should maintain the same preference; however, in practice, team members may have different estimates of the “shared” preference (i.e., misalignment of mental models) due to ineffective team communications, lack of expertise, or a variety of execution-time factors.

To infer and \emph{predict} misalignment of mental models, we provide a Bayesian approach that leverages prior knowledge about the collaborative task and data of the team members’ task execution (namely, actions and observable states of the task).
As an exemplary application, we demonstrate our approach using two simulated collaboration scenarios, inspired by teamwork during cardiac surgery procedures (e.g., CABG, SAVR, etc): namely (a) protamine administration to reverse heparin and (b) surgical tool handovers.
In the experiments conducted on simulated tasks, our approach could infer model misalignment with over 75\% recall.
Encouraged by these results, in our ongoing work, we are extending our approach to enable its use in a simulated operating room (OR).
The paper concludes with a brief discussion of this ongoing and future work, with emphasis on the computational challenges.

\section{Representing Collaborative Tasks}
\label{sec:task-model}

A mathematical representation of teamwork is necessary to enable automated reasoning of team behavior.
Motivated by the variety of teaming contexts, research on human teaming, multi-agent systems and human-robot collaboration has led to several formalism to represent teamwork \cite{grosz1999evolution, schurr2004steam, amir2016mip}.
Here, we utilize a multi-agent and partially observable variant of MDPs, to represent goal-oriented collaborative tasks of interest.

MDPs provide a framework to represent sequential decision-making tasks \cite{oliehoek2016concise}.
They are specified by a set of states $s \in \mathcal{S}$, which represent the task context; a set of actions $a \in \mathcal{A}$, which represent the actions that can be taken while performing the task; a transition model $T: \mathcal{S} \times \mathcal{A} \times \mathcal{S} \rightarrow [0,1]$, which specifies the distribution over the next task state given current state and action; and a reward function $R: \mathcal{S} \rightarrow \Re$, which specifies the reward of reaching a state and can encode task goals.
As MDPs can encode stochastic outcomes and goals, they provide a useful framework to represent real-world sequential tasks.

MDPs, however, represent tasks with one decision-maker.
To represent teams, we consider its multi-agent variants, where the action represents joint action of a team.
Specifically, for a team with $n$ members, the joint action is represented as $a = [a_1, a_2, \cdots, a_n]$, thereby enabling modeling of multiple decision-makers.
The task progress (through the transition model), now, depends on the joint action (i.e., action of each team member).
The behavior of a team, for a fully observable multi-agent MDP, can be specified using $n$ policies, where $\pi_i(a_i | s)$ denotes the policy of the $i$-th team member.

In real-world tasks, the state may not be fully observable. Especially for humans, their preferences or (partial) understanding of the task may influence their behavior.
Similarly, for human teams, situation awareness may influence each team member’s policy.
Thus, we augment the task state, with a latent feature $x \in X$, which denotes the team member’s latent preferences regarding the task.
In general, the latent feature may evolve during the task; however, here, we limit our scope to time-invariant latent states.
Thus, the collaborative task can be described using the tuple $(\mathcal{S}, \mathcal{X}, \mathcal{A}, T, R)$.
Next, we instantiate this task model for two scenarios inspired by teamwork in the OR.

\subsection{Protamine Administration}
\label{sec:protamine_description}

Cardiac surgery is performed by a team of surgeons, anesthesiologists, perfusionists, and nurses (as shown in Figs.~\ref{fig:teamwork-or}-\ref{fig.tooldeliveryenv}); in academic medical centers, a subset of the team can be trainees (e.g. residents, fellows, students) with limited task experience.
After successful weaning of the patient from the cardioplumonary bypass machine, alongside removal of the venous and arterial cannulas, protamine needs to be administered to reverse the anticoagulant effect of heparin and restore normal blood coagulation.
The resident anesthesiologist (RA) administers protamine after receiving a verbal request from the attending surgeon (AS), after which the surgeon begins removing the cannulas.
One caveat is that a patient may be allergic to protamine leading to a life-threatening “protamine reaction” syndrome; hence, protamine needs to be administered intravenously incrementally over 5-10 minutes and not as a single intravenous bolus.

Administration of protamine as a bolus (i.e., all at once) can lead to a protamine reaction with refractory vasoplegia unless action is not taken quickly and efficiently by the team; however, due to limited prior experience, an RA can have an incorrect mental model of the team's strategy (incremental vs. bolus administration).
In a prior documented case in the literature, protamine administration was conducted by the RA improperly as a bolus, despite the oversight of the attending anesthesiologist \cite{zenati2019first}.
Meanwhile, physical barriers including the sterile drape separating the RA from the AS preclude explicit awareness of such an error by the team unless verbally communicated.
Thus, misalignment of mental models may occur if the team is not proceeding through the surgical steps as expected.
If an AI Coach can infer this lack of shared mental models (between the AS and RA), it can help prevent associated adverse outcomes.

Thus, as our first task, we represent the collaboration between the AS and RA during protamine administration using the task model.
The goal of the team (encoded as $R$) is to safely administer protamine and successfully remove cannulas while avoiding any adverse outcomes.
The observable component of the task state $s \in \mathcal{S}$ is defined by the following features: protamine administration phase ($s_1$, a Boolean variable denoting whether the surgical workflow is in the protamine administration phase), status of protamine dosage ($s_2$, indicating the percentage amount of protamine administered), number of cannulas removed ($s_3$), and patient state ($s_4$, indicating the patient state, as measured by the vitals and categorized as nominal, allergic, and adverse).

In addition to the observable features, which can be measured using sensors and instrumentation in the operating room \cite{kennedy2020computer, kennedy2020sensors}, the team behavior depends on the proper understanding of the task.
In particular, the team may have one of two mental states regarding protamine administration: bolus or incremental (especially because heparin is always given as a bolus).
These mental states correspond to the latent state $x \in \mathcal{X}$ in the task model, as they cannot be measured by a physiological sensor.
The surgeon is modeled to have the following actions $a_1 \in \mathcal{A}_1$: request protamine, remove cannula, and No-op.
Similarly, the anesthesiologist can take the following actions $a_2 \in \mathcal{A}_2$: administer incremental dosage, administer bolus, communicate, and No-op.

The transition model $T$ represents the effect of team members' actions on the task state.
The task begins prior to the protamine administration phase $(s_1 = 0)$, and transitions to the protamine administration phase $(s_1 = 1)$ after the surgeon's `request' action.
The `remove cannula' action updates the status of cannula removal $(s_3)$ deterministically.
Similarly, the protamine administration actions update the status of protamine administration $(s_2)$; specifically, `bolus' changes $s_2 = 100\%$, where `incremental dosage' increments $s_2$ first by a test dosage and, subsequently, by $25\%$.
The incremental dosage leads to an allergic reaction with $0.01$ probability.
Similarly, bolus administration of protamine leads to an adverse reaction with $0.8$ probability.
The No-op and `communicate' actions do not change the task state; however, we posit that a more engaged resident (with the correct task understanding) is likely to communicate more often.
The scenario terminates after the goal is achieved or when the patient exhibits adverse or allergic reaction, after which the surgical team adopts specific protocols to restore the patient's stability.

\subsection{Surgical Tool Delivery}
\label{sec:tooldelivery_description}

\begin{figure}
\centerline{%
    \includegraphics[width=0.9\columnwidth]{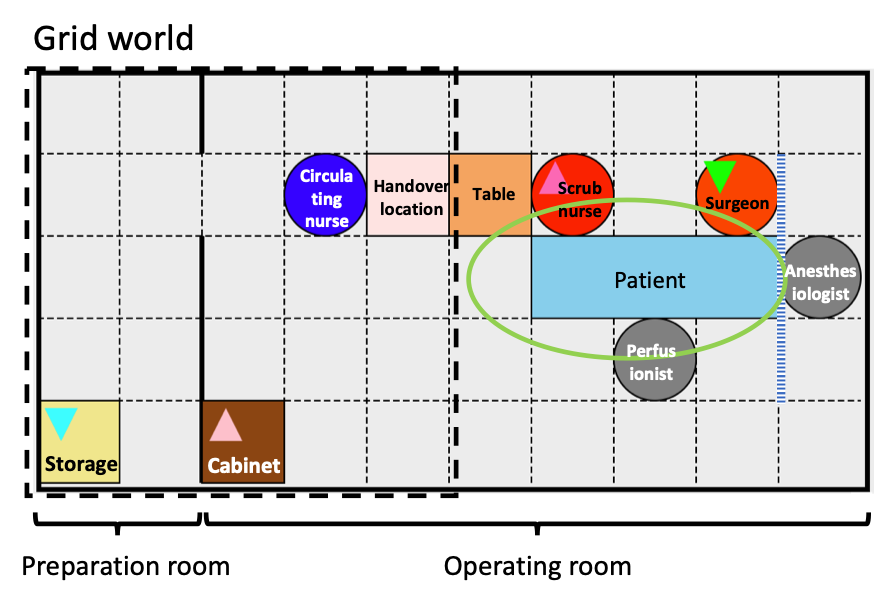}
}
\caption{The tool delivery environment. Circulating nurse (CN) cannot enter the sterilized area (circled with green line), while the scrub nurse (SN) cannot retrieve unsterilized items. Hence, during the course of the surgery, SN may solicit CN's help to acquire items from the cabinet or storage area.}
\label{fig.tooldeliveryenv}
\vspace{-1em}
\end{figure}

As the second task, we model the handovers (between scrub and circulating nurses) encountered in surgery.
To model this task, we consider a grid-world representation of the operating area (Fig.~\ref{fig.tooldeliveryenv}) and focus on the sub-team of surgeon, scrub nurse (SN), and circulating nurse (CN).
During the preoperational stage, required surgical tools are prepared and placed in the sterile area next to the operating table.
However, every so often, an additional tool (or item, such as sutures) may be required to be delivered from outside the sterile area.
In such situations, SN may ask the CN to deliver the requisite tool.
Incorrect tool delivery, due to lack of shared mental models arising from ineffective communication of the surgeon’s preferences, can delay the surgery and lead to preventable harm in time-critical situations.
In such situations, an AI Coach can help detect model misalignment and mitigate delivery of incorrect tools.

In our simulated environment, the CN can move in a 5-by-5 grid world (Fig.~\ref{fig.tooldeliveryenv}), while the surgeon and SN are limited to the sterile area.
The task begins with the SN having sterile sutures and scalpels.
Additional sutures are located in the cabinet in the operating room, while scalpels in the storage area (in an adjacent room).
During the task, SN may require additional sutures or replacement scalpel (i.e, $\mathcal{X} = \{\textit{Sutures}, \textit{Scalpel}\}$), and communicate this requirement to CN.
If the CN misunderstands the SN's request, the team members will have misaligned mental models $\mathcal{X}$.

The requirement of additional items depends on observable features of the surgery.
For instance, if the scalpel falls off of the sterile field and become contaminated, the SN is more likely to request replacement scalpel.
Similarly, if the next step in surgery involves suturing, the SN is more likely to request sutures.
Thus, in addition to the latent feature $x$, we model the task using the following observable features $s \in \mathcal{S}$: tool positions, patient status, request status, and position of the circulating nurse.
The goal of the collaboration scenario (encoded as $R$) is for the team to have access to the correct tool.
Transition model encodes the effect of team member's action on the task state.
For instance, tool position changes if they are picked and delivered by the CN.

\section{Online Inference of Model Alignment}
\label{sec:bayesian_approach}

As demonstrated above, the task model can be used to represent a variety of collaborative tasks.
In these tasks, both observable and latent features influence the team.
While observable features $s$ are shared among the team, the value of latent feature $x$ can differ across teammates.
For instance, during tool delivery, the circulating nurse might retrieve the incorrect tool, due to an incorrect assumption about the requested tool (i.e., the latent feature of the task).
Such misalignment can lead to lack of shared situation awareness, poor collaboration, and preventable harm.
Hence, towards the goal of developing an AI Coach that can mitigate preventable harm, in this section, we provide an algorithm to infer model alignment (or lack thereof) through observation of team's task execution.

\subsection{Problem Statement}
To arrive at the algorithm, we first provide a mathematical description of the problem statement.
For a given task specification, we assume that each team member as well as the AI Coach can observe (through sensors) the task's observable features and, thus, maintains a shared understanding of the state component $s$.
However, each team member can potentially maintain a different understanding regarding the latent feature $x$, where the estimate of $i$-th team member is denoted as $\hat{x}_i$.
Further, the policy of the $i$-th team member depends on both latent and observable features, and is denoted as $\pi_i(a_i | s, \hat{x}_i)$.
In a team with shared understanding, all team members will maintain the same estimate of the latent feature (i.e., $\hat{x}_1 = \hat{x}_2 = \cdots = \hat{x}_n$).
However, if the estimates differ, the team members will have misaligned mental models.

The AI Coach seeks to infer this model misalignment given the task model $(\mathcal{S}, \mathcal{X}, \mathcal{A}, T, R)$, team members' policies $(\pi_1, \pi_2, \cdots, \pi_n)$, and data of team's task execution (i.e., $s,a$-sequences).
The task execution sequences are represented as $\tau = (s^0, a^0, s^1, a^1, ... s^k, a^k)$, where the subscript denote the sequence indices, $s^0$ denotes the initial task state, and $k$ denotes the sequence length.
Note that only observable state features can be sensed, thus the task execution data excludes latent feature of the task state $(x)$.

\subsection{Inference Algorithm}
We adopt a Bayesian approach to infer model misalignment.
To estimate the quantity of interest $(x)$, in addition to data, Bayesian algorithms require specification of prior probability $p(x)$ and likelihood model $p(\text{data} | x)$.
Here, we utilize the task and policy specifications to arrive at the likelihood model, while noting that, in general, its specification is non-trivial.
In our ongoing work, we are actively developing approaches to learn these models from training data, for the case where team members' policies are difficult to specify.

In our case, the Bayesian algorithm seeks to infer the latent state $\hat{x}^i$ for each member of the team.
Let us denote this joint estimate as $\hat{\mathbf{x}} = (\hat{x}_1, \hat{x}_2, ... \hat{x}_n)$.
Given the data, the posterior probability is computed as follows,
\begin{align}
    p( \hat{\mathbf{x}} | \tau)
    &\propto p(\tau | \hat{\mathbf{x}}) p(\hat{\mathbf{x}}) \nonumber
    = p(s^0, \mathbf{a}^0, s^1, \cdots, s^{t}| \hat{\mathbf{x}}) p(\hat{\mathbf{x}}) \nonumber \\
    &= p(\hat{\mathbf{x}}) p(s^0) \prod_{j=0}^{k-1} T(s^{j+1} | \mathbf{a}^{j}, s^{j})  P(\mathbf{a}^{j} | s^{j}, \hat{\mathbf{x}}) \nonumber \\
    &\propto \prod_{i=1}^{n} \left( p(\hat{x}_i) \prod_{j=0}^{k-1} \pi_i (a^{j}_{i} | s^{j}, \hat{x}_i) \right) \label{eq:x-posterior-probability},
\end{align}
where, we assume that the prior probabilities and policies corresponding to each team member are independent, i.e.,
\begin{align}
p(\hat{\mathbf{x}}) &= p(\hat{x}_1) p(\hat{x}_2) \cdots p(\hat{x}_n) \\
P(\mathbf{a} | s, \hat{\mathbf{x}}) &= \pi_1 (a_{1} | s, \hat{x}_1) \pi_2 (a_{2} | s, \hat{x}_2) \cdots \pi_n (a_{n} | s, \hat{x}_n).
\end{align}

Given \ref{eq:x-posterior-probability}, the latent state is inferred as the maximum posteriori estimate.
In the inferred latent state of each agent are not identical, then the algorithm reports a model misalignment.

\section{Experiments}
\label{sec:experiments}

To evaluate the inference approach, we utilize the collaboration scenarios described in Sec.~\ref{sec:task-model}.
For each scenario, we implement the Markovian task model (detailed in Sec.~\ref{sec:protamine_description}-\ref{sec:tooldelivery_description}), specify ground truth policies of the team members, and generate synthetic data of task execution using the task and policy specifications.
Execution sequences are created by first assigning latent states $\hat{x}_i$ to the team member and, then iteratively, (a) sampling team members' action $a$ using their policy, latent state $\hat{x}_i$, and task state $s$, and (b) sampling the next state $s'$ in the sequence using the transition model $T(s' | s, a)$, until the task termination criteria is reached.

\subsection{Protamine Administration}

We generate $300$ task sequences for the simulated protamine administration task, where model misalignment could occur due to incorrect task understanding.
In our simulations, the AS always expects incremental protamine administration, while the RA may (incorrectly) administer it as a bolus with probability $0.5$.
We describe the team member's policies next.

The scenario begins prior to protamine administration phase, during which the RA may communicate (e.g., ask for supervision, provide updates) with AS.
We model that an RA which communicates more often is less likely to have an incorrect understanding of the task (i.e., misaligned mental model).
During the task, the AS initiates the protamine administration phase through a verbal communication, after which protamine is administered bolus or incrementally by the RA based on their task understanding $\hat{x}$.
Cannula removal is interleaved with the protamine administration. 
The AI Coach can sense the observable state $s$, team's actions $a$, and seeks to infer $\hat{\mathbf{x}}$.
Among the $300$ task sequences generated, in $155$ the team exhibits model misalignment.
We evaluate the inference performance with both full and partial task sequences.

\textbf{Post-hoc performance} As the AI coach can monitor the amount of protamine administered, with full task sequences, it can always infer model misalignment accurately.
While post-hoc inference of misalignment does not prevent the anesthesiologist from performing the erroneous action (bolus), it can help identify near-miss events (i.e., where the incorrect action did not lead to an adverse outcome; $31$ out of $300$ in our synthetic data) and help the team achieve a shared mental model for subsequent surgeries.

\textbf{Execution-time performance} Since the algorithm does not impose any condition on the length of sequence, it can also provide an estimate during the task (i.e., given partial task execution).
For the current task, we evaluate this capability for when the surgeon requests protamine (i.e., before a potential error).
Even with a partial task sequence, the algorithm results in overall estimation accuracy of 66.3\%.
In life-critical tasks, where the goal of AI Coach is to mitigate adverse outcomes, false alarms are less critical.
Among the $155$ sequences with misaligned mental models, the algorithm could predict model misalignment in $119$ cases, i.e., exhibiting 76.8\% recall.

\subsection{Surgical Tool Delivery}

For the tool delivery task, we generate $300$ synthetic task sequences and $271$ samples among them include the situation where a tool was requested.
The task sequences include the positions of the tools and CN, patient status (i.e., whether an incision has been made or not), item request status (i.e., whether SN has requested an item from CN), and actions of the team members.
While the AI Coach can sense when an item request is made, to reflecting sensing capabilities in the OR, it cannot detect which item was requested in our simulations.
Thus, the inference algorithm needs to infer the latent states (i.e., tool being requested, $\hat{x}$) for both SN and CN, to estimate model alignment.
Similar to the protamine case, we evaluate both the post-hoc and execution-time performance.
On average, the full sequences are $\approx 30$ steps; while the partial sequences include $7$ steps (2 steps before the request and 5 steps after the request).
The inference algorithm results in 75.9\% and 98.5\% recall with partial and full sequences, respectively.

\section{Discussions}
In this paper, we propose the problem of inferring mental model alignment in collaborative tasks and provide an execution-time approach to infer model alignment using observable features of team behavior.
Through computer simulations of surgery-inspired collaborative scenarios, we provide proof-of-concept results that demonstrate that the inference of model alignment is computationally feasible and can help mitigate preventable harm.
For instance, in the protamine administration scenario, the proposed approach could predict model misalignment (and associated errors) 76.8\% given task model, policy specifications, and partial task sequences.

Further, even in the challenging tool delivery task where multiple strategies may be valid (e.g., request of either tool is valid in the tool delivery domain), the inference algorithm could infer model misalignment with 75.9\% recall with partial task sequences (i.e., before the tool is delivered).
In addition to mitigating preventable harm, thus, this execution-time inference of can also help improve team's task performance.
For instance, an AI Coach that provides interventions based on inferred model misalignment, can help in the early detection of incorrect tool delivery; note that, without an AI Coach, the team would realize their error only after it is made.

Encouraged by these results, we are working towards the vision of realizing an AI Coach for human teams by relaxing the requirement of model specification and addressing associated sensing challenges.
For instance, the proposed approach assumes accurate specification of team members' policies and complete observablity of their actions, both of which might be difficult to meet in practice.
Hence, we are exploring learning-based approaches to arrive at team policies in presence of latent states \cite{osa2018algorithmic, unhelkar2019learning}.
To address the challenges associated with state and action observability, the development of an AI Coach would be greatly enhanced by nuanced surgical tool detection and people tracking methodology.
The possibilities of incorporating these approaches into the OR are steadily increasing as surgical data science takes hold in surgical environments \cite{kennedy2020computer}.
Additionally, robust and sensitive psychophysiological sensors equipped to team members may provide insight into impending mental model misalignments and function to provide automated feedback on team members' cognitive states.



\bibliographystyle{IEEEtran}
\bibliography{main}
\end{document}